\documentclass[letterpaper, 10 pt, conference]{ieeeconf}
\usepackage{cite}
\usepackage{amsmath,amssymb,amsfonts}
\usepackage{algorithmic}
\usepackage{graphicx}
\usepackage{textcomp}
\usepackage{multirow}
\usepackage{makecell}
\usepackage{subcaption}
\usepackage{bm}
\usepackage{svg}
\usepackage[utf8]{inputenc}
\usepackage{microtype}
\usepackage{booktabs}
\usepackage{url}
\usepackage{float}
\usepackage[top=1.91cm, left=1.91cm, right=1.91cm, bottom=1.91cm]{geometry}
\usepackage{tikz}
\usepackage{pgfplots}
\pgfplotsset{compat=1.18}
\usepackage{algorithm}
\usepackage{caption}

\usepackage{hyperref}
\usepackage{xcolor}
\hypersetup{
    colorlinks=true,
    linkcolor=red,
    filecolor=red,      
    urlcolor=black,
    linktoc=all,
    citecolor=green,
}

\IEEEoverridecommandlockouts

\title{\LARGE \bf Correspondence-Free Pose Estimation with Patterns: A Unified Approach for Multi-Dimensional Vision}
\author{ Quan Quan, and Dun Dai\thanks{%
Quan Quan, and Dun Dai are with the School of Automation Science and Electrical Engineering, Beihang University, Beijing 100191, China
(e-mail: \{qq\_buaa, d\_dai\_3\}@buaa.edu.cn.)}}

\begin{document}
\maketitle
\thispagestyle{empty}
\pagestyle{empty}

\begin{abstract}
6D pose estimation is a central problem in robot vision. Compared with pose estimation based on point correspondences or its robust versions, correspondence-free methods are often more flexible. However, existing correspondence-free methods often rely on feature representation alignment or end-to-end regression. For such a purpose, a new correspondence-free pose estimation method and its practical algorithms are proposed, whose key idea is the elimination of unknowns by process of addition to separate the pose estimation from correspondence. By taking the considered point sets as patterns, feature functions used to describe these patterns are introduced to establish a sufficient number of equations for optimization. The proposed method is applicable to nonlinear transformations such as perspective projection and can cover various pose estimations from 3D-to-3D points, 3D-to-2D points, and 2D-to-2D points. Experimental results on both simulation and actual data are presented to demonstrate the effectiveness of the proposed method.
\end{abstract}

\section{Introduction}
In recent years, mobile intelligent robots have attracted increasing
attention in academic research and industrial applications
\cite{Q.Quan(2017)}. They play an important role in some scenarios such as search and rescue, goods delivery, surveillance, and agricultural applications. An accurate localization algorithm is a basis and prerequisite for these tasks. Therefore, how to achieve efficient and autonomous pose (3D-motion) estimation is a hot topic for mobile robots, especially in some environments where the Global Positioning System (GPS) signal is not available.

Traditional hand-crafted feature detectors and descriptors (such as SIFT\cite{SIFT2004}, ORB\cite{ORB}, and SURF\cite{surf}), where correspondences are established by filtering descriptor pair distances iteratively. When comes to the 3D-3D case, Iterative Closest Points (ICP)\cite{ICP} algorithm and its variants use hand-crafted geometric proprieties like coordinates and surfaces to construct the correspondence. These early methods rely on improving the accuracy of local feature extraction to effectively enhance matching quality. Feature-based methods have been
widely used in various mobile robot localization \cite{FXiao(2018)}, \cite%
{EWestman(2018)}, which can reliably solve the positioning problem. However,
they usually rely on the correctness and efficiency of feature extraction
and matching, which may fail in poorly textured environments or
situations with defocus and motion blur. Without a good pose initialization
or restrictive assumptions, it is hard to find the optimal solution\cite{McFadyen(2016)}. 
\begin{figure}[t]
\centerline{\includegraphics[
		scale=0.8]{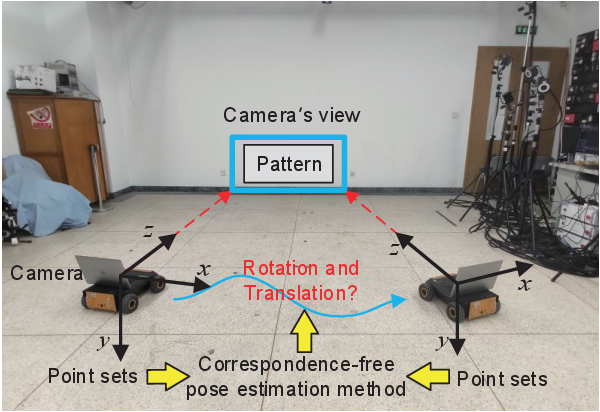}}
	\vspace*{-5pt}
\caption{Indoor experimental setup for correspondence-free pose estimation using wall patterns. The camera follows a trajectory while maintaining sufficient pattern visibility within its field of view.}
\label{fig1}
\end{figure}
In the deep learning era, learning-based methods have exhibited extraordinary capabilities in pattern recognition. Recent advances in this field can be broadly classified into two directions. One approach focuses on devising methods to extract more distinctive features, thus moving beyond traditional handcrafted techniques (e.g., \cite{Verdie2015, magicpoint}). The other strategy centers on directly enhancing correspondence accuracy (e.g., \cite{deng20193d, Pais2020, Zhou2024, OANET, superglue, lindenberger2023}). Both directions have substantially boosted pose estimation performance through pre-training and generalizing specific components within the relative pose estimation pipeline.

In contrast, correspondence-free methods eliminate the need for explicit one-to-one matching. One prominent category within this framework involves learning global representations of sparse features, where the embedded representations are designed to be approximately equivariant under rotational and translational transformations. Another end-to-end approach directly regresses the relative pose from paired images, offering a straightforward solution that preserves scale information.

Departing from established approaches, this paper proposes a complementary method to the feature-based method because it can obtain a good pose initialization for feature-based methods to reduce the matching time. More importantly, the proposed method is also new and independent of existing correspondence-free methods\cite{cohen2016, deng2021,zhu2022} that often rely on aligning global representation or the methods \cite{Shavit2021, Rockwell2022} that predict the pose in an end-to-end way. To formulate the problem, we consider relative pose estimation based on two sets of points with unknown correspondences. First, the relation between each corresponding point pair is expressed as an equation, with each element of the pair appearing on opposite sides. This formulation applies to pose estimation problems involving 2D-to-2D, 3D-to-3D, and 3D-to-2D correspondences. Subsequently, by summing the equations corresponding to all point correspondences, we obtain aggregated correspondence-free equations since the order of the summation is irrelevant. Similarly, by considering the point sets as patterns, feature functions are introduced to generate a sufficient number of equations that are correspondence-independent and handle cases involving mismatches. Ultimately, the pose estimation problem is reformulated as a set of optimization problems that do not rely on explicit correspondences. 

The main contributions of this paper are as follows:

\begin{itemize}
\item  We present a novel, efficient, and robust correspondence-free pose estimation method. Instead of relying on refined one-to-one correspondences, our approach determines the relative pose globally by formulating and solving a set of optimization equations.

\item  Different from the correspondence-free method which only works for a single case, the proposed method is flexible and applicable to 3D-to-3D, 3D-to-2D, and 2D-to-2D points. 

\item  Extensive experiments under challenging conditions, including noise, mismatches, occlusions, and outliers. Demonstrate the practicality and reliability of the proposed approach.
\end{itemize}

\section{Problem Formulation}

Assume that we have a pattern in the sight of a moving robot with a camera, as shown in 
Fig.~\ref{fig1}. In fact, the image-based visual servo easily keeps the whole pattern in the field of view of the camera \cite{Yang2020}. Then, the camera points to the pattern, and two point sets are obtained between consecutive images, defined as
{\setlength\abovedisplayskip{2.5pt}
	\setlength\belowdisplayskip{2.5pt}
\begin{align}
\mathcal{P}& =\{p_{k}\in \mathbb{R}^{n_{p}},k=1,\cdots ,N\} \\
\mathcal{Q}& =\{q_{k}\in \mathbb{R}^{n_{q}},k=1,\cdots ,N\}
\end{align}}%
where $N=\left \vert \mathcal{P}\right \vert =\left \vert \mathcal{Q}\right
\vert.$ The symbol $\left \vert \mathcal{S}\right \vert $ denotes the
number of elements of the point set $\mathcal{S}$. To simplify the derivation,  
we assume that the numbers of the two point sets are the same. The general situation with different numbers of points is further discussed in Section %
\uppercase \expandafter{\romannumeral3} as follows. For the point sets, 
there exist functions $h:{\mathbb{R}^{n_{p}}}\times {\mathbb{R}^{m}}%
\rightarrow {\mathbb{R}^{n}}$ such that 
\begin{equation}
h\left( p_{k},\theta \right) =q_{c\left( k\right) }, k=1,\cdots ,N
\label{mappingsim}
\end{equation}%
where $\theta \in \mathbb{R}^{m}$ is an unknown parameter vector and $%
c:\left \{ 1,\cdots ,N\right \} \rightarrow \left \{ 1,\cdots ,N\right \} $
is an unknown bijection, called the correspondence function. The $q_{c\left(
k\right) }$ represents the matching point $p_{k}$ through the function $c$.

The objective is to design a correspondence-free method to estimate $\theta $
based on sets$\  \mathcal{P},\mathcal{Q}$ and the function $h$. For
simplicity, this problem is called the \textbf{Estimation Problem} on $%
\left( \mathcal{P},\mathcal{Q},h,\theta \right) .$ In the following, some existing
pose estimation problems are simply formulated in the form of (\ref%
{mappingsim}).

\begin{itemize}

\item \textbf{3D-to-3D Example}. The relation
of 3D-to-3D points can be formulated as (\ref{mappingsim}) directly with 
\begin{equation}
\left \{ 
\begin{array}{l}
\mathcal{P}=\left \{ {M}_{k}\in \mathbb{R}^{3},k=1,\cdots ,N\right \}  \\ 
\mathcal{Q}=\left \{ {M}_{k}^{\prime }\in \mathbb{R}^{3},k=1,\cdots
,N\right \}  \\ 
h\left( {M}_{k},\theta \right) =RM_{k}+{T}{.}%
\end{array}%
\right. 
\end{equation}

\item \textbf{3D-to-2D Example}.  For the pose estimation problem from
3D-to-2D points, $M_{k}$ and corresponding homogeneous image coordinate $m_{c\left( k\right)}^{\prime }$ (recovered from pixel coordinate with known intrinsic matrix) are known. The corresponding depths $z_{c\left( k\right) }^{\prime }$ is unknown. Fortunately, since $\frac{%
M_{c(k)}^{\prime }}{\left \Vert M_{c(k)}^{\prime }\right \Vert }=\frac{%
m_{c(k)}^{\prime }}{\left \Vert m_{c(k)}^{\prime }\right \Vert },$ we can
modify $RM_k+T=M'_k$ to be 
\begin{equation}
\frac{{R}M_{k}+T}{\left \Vert {R}M_{k}+T\right \Vert }=\frac{m_{c\left(
k\right) }^{\prime }}{\left \Vert m_{c\left( k\right) }^{\prime }\right
\Vert }.
\end{equation}%
As a result, the relation of 3D-to-2D points can be formulated in the form of (\ref%
{mappingsim}) with 
\begin{equation}
\left \{ 
\begin{array}{l}
\mathcal{P}=\left \{ M_{k}\in \mathbb{R}^{3},k=1,\cdots ,N\right \} \\ 
\mathcal{Q}=\left \{ \frac{m_{k}^{\prime }}{\left \Vert m_{k}^{\prime
}\right \Vert }\in \mathbb{R}^{3},k=1,\cdots ,N\right \} \\ 
h\left( M_{k},\theta \right) =\frac{RM_{k}+T}{\left \Vert RM_{k}+T\right
\Vert }{.}%
\end{array}%
\right. \label{3dto2d}
\end{equation}

\item \textbf{2D-to-2D Example. }The relation of 2D-to-2D points is also compatible with our method. First, all the
reference points\ are considered to lie in a plane $n^{T}M_{k}-d=0,$ $%
k=1,\cdots ,N,$ where $n\in \mathbb{R}^{3}$ and $d\in \mathbb{R}.$ In this
case, the $k$th points in the two camera coordinates are related by \cite[%
p.327]{Hartley(2003)}%
\begin{equation}
\frac{z_{k}}{z_{c\left( k\right) }^{\prime }}\left( R+\frac{1}{d}%
Tn^{T}\right) m_{k}=m_{c\left( k\right) }^{\prime }.  \label{plane}
\end{equation}%
By eliminating the unknown term ${z_{k}}/{z_{c\left( k\right)
}^{\prime }},$ the equation (\ref{plane}) is further written as%
\begin{equation}
\frac{\left( R+\frac{1}{d}Tn^{T}\right) m_{k}}{\left \Vert \left( R+\frac{1}{d%
}Tn^{T}\right) m_{k}\right \Vert }=\frac{m_{c\left( k\right) }^{\prime }}{%
\left \Vert m_{c\left( k\right) }^{\prime }\right \Vert }.
\end{equation}%

Consequently, the relation of arbitrary 2D-to-2D points can be formulated as 
\begin{equation}
\left \{ 
\begin{array}{l}
\mathcal{P}=\left \{ m_{k}\in \mathbb{R}^{3},k=1,\cdots ,N\right \}  \\ 
\mathcal{Q}=\left \{ m_{k}^{^{\prime }}\in \mathbb{R}^{3},k=1,\cdots
,N\right \}  \\ 
h\left( m_{k},\theta \right) =\frac{\left[ T\right] _{\times }Rm_{k}}{%
\left \Vert \left[ T\right] _{\times }Rm_{k}\right \Vert } \\ 
g\left( m_{k}^{\prime },\theta \right) =\frac{\left[ T\right] _{\times
}m_{k}^{\prime }}{\left \Vert \left[ T\right] _{\times }m_{k}^{\prime
}\right \Vert }{.}%
\end{array}%
\right. 
\label{2dto2d}
\end{equation}%
where $[\mathbf{\cdot}]_\times$ denotes the skew-symmetric matrix form. The introduction of skew-symmetric matrix\cite[%
p.108]{Q.Quan(2017)} transforms the cross-product operation into an equivalent matrix multiplication form, which facilitates the subsequent pose optimization process and the computation of Jacobian matrices. We should notice that $T=0$ cannot be a solution. which means that the additional restriction on $T$%
, such as $\left \Vert T\right \Vert =1,$ is necessary.

\end{itemize}

\section{Correspondence-Free Pose Estimation}

To better understand the idea, a motivation example is given first, where the parameter estimation and the
correspondence estimation is decoupled so that the proposed method requires
less computation. 

\subsection{A Motivation Example}

Assume that we have two sets
\begin{align}
\mathcal{P}& =\left \{ p_{1}=1,p_{2}=2,p_{3}=3,p_{4}=4\right \} \\
\mathcal{Q}& =\left \{ q_{1}=2,q_{2}=3,q_{3}=1,q_{4}=4\right \} .
\end{align}
There exists a function $\theta p_{k}=q_{c\left( k\right) }$ mapping $%
\mathcal{P}$ to $\mathcal{Q}$, where both the parameter $\theta \in \mathbb{R%
}$ and the correspondence function $c\left( \cdot \right) $ are unknown. The
objective is to estimate $\theta.$ Although $\theta p_{k}=q_{c\left(
k\right) }$ holds, $\theta ={q_{c\left( k\right) }}\left/ {p_{k}}\right. $
cannot help as the correspondence $\left( p_{k},q_{c\left( k\right) }\right) 
$ is unknown. In most literature, the recovery of parameters requires
correspondence between the two sets. This will be more and more difficult
as the number of correspondences increases. In the following, we will
introduce a new idea to handle the estimation problem without
correspondence.

First, we add all equations together as $\theta 
\underset{k=1}{\overset{4}{\sum }}p_{k}=\underset{k=1}{\overset{4}{\sum }}%
q_{c\left( k\right) }.$ It is noticed that the value $\underset{k=1}{\overset%
{4}{\sum }}q_{c\left( k\right) }$ is independent of the correspondence
function $c\left( \cdot \right) ,$ i.e., $\underset{k=1}{\overset{4}{\sum }}%
q_{c\left( k\right) }=\underset{k=1}{\overset{4}{\sum }}q_{k}$ for an
arbitrary correspondence function $c\left( \cdot \right) $. Then%
\begin{equation}
\theta =\underset{k=1}{\overset{4}{\sum }}q_{c\left( k\right) }\left/ 
\underset{k=1}{\overset{4}{\sum }}p_{k}\right. =\underset{k=1}{\overset{4}{%
\sum }}q_{k}\left/ \underset{k=1}{\overset{4}{\sum }}p_{k}\right. =1.
\end{equation}

The key idea is that\emph{\ changing the order of the correspondence does
not change the sum}. By reducing four equations to one, we can estimate $%
\theta.$ With the estimate $\theta ,$ we can easily obtain the
correspondence function as%
\vspace*{-5pt}
\begin{equation}
c\left( 1\right) =3,c\left( 2\right) =1,c\left( 3\right) =2,c\left( 4\right)
=4.
\end{equation}
\vspace*{-18pt}

Compared with simultaneous parameter and correspondence estimation, the
dimension of the parameter space is very low. It is easy to see that the
computational complexity of the method for the simple example is only $%
O\left( N\right) .$ Then, we extended the idea to the proposed pose estimation problem.

\subsection{Basic Method}

Under the considered estimation problem, the dimension of the parameter space
is $m$ so that at least $m$ equations independent of the correspondence
function $c\left( \cdot \right) $ are required to obtain the estimate of $%
\theta \in \mathbb{R}^{m}$. For such a purpose, we select a class of
independent nonlinear functions $f_{i}:\mathbb{R}^{n}\rightarrow \mathbb{R},$
$i=1,\cdots ,L\geq m$ to establish equations according to (\ref{mappingsim}) as
follows%
\vspace*{-5pt}
\begin{equation}
\frac{1}{N}\underset{k=1}{\overset{N}{\sum }}f_{i}\left( h\left(
p_{k},\theta \right) \right) =\frac{1}{N}\underset{k=1}{\overset{N}{\sum }}%
f_{i}\left( q_{c\left( k\right) }\right) .
\end{equation}

Denote $x=\left[ 
\begin{array}{ccc}
x_{1} & \cdots & x_{n}%
\end{array}%
\right] ^{T}.$ The simplest feature functions are $f_{1}\left( x\right)
=x_{1},\cdots ,f_{n}\left( x\right) =x_{n}$ which result in%
\begin{equation}
\frac{1}{N}\underset{k=1}{\overset{N}{\sum }}h\left( p_{k},\theta \right) =%
\frac{1}{N}\underset{k=1}{\overset{N}{\sum }}q_{k}.
\end{equation}%

The average is a commonly-used feature function. The equations $
\frac{1}{N}\underset{k=1}{\overset{N}{\sum }}f_{i}\left( q_{c\left(k\right)} \right) =\frac{1}{N}\underset{k=1}{\overset{N}{%
\sum }}f_{i}\left( q_{k}\right)$ 
are independent of the correspondence function $c\left( \cdot \right) ,$ $%
i=1,\cdots ,L.$ Consequently, for any arbitrary correspondence function $%
c\left( \cdot \right) $, we have%
\begin{equation}
\frac{1}{N}\underset{k=1}{\overset{N}{\sum }}f_{i}\left( h\left(
p_{k},\theta \right) \right) =\frac{1}{N}\underset{k=1}{\overset{N}{\sum }}%
f_{i}\left( q_{k}\right) ,i=1,\cdots ,L.  \label{LumpEqu1}
\end{equation}%

Based on the equations (\ref{LumpEqu1}), the parameter $\theta $ can be
obtained by solving the following optimization problem%
\begin{equation}
\underset{\theta }{\min }\left \Vert F_{p}\left( p,\theta \right)
-F_{q}\left( q\right) \right \Vert ^{2}  \label{optimization0}
\end{equation}%
where%
\begin{align}
p& =\left[ p_{k}\right] _{N\times 1}\in \mathbb{R}^{n_{p}N}, \\
q& =\left[ q_{k}%
\right] _{N\times 1}\in \mathbb{R}^{n_{q}N}, \\
F_{p}\left( p,\theta \right) & =\left[ \frac{1}{N}\underset{k=1}{\overset{N}{%
\sum }}f_{i}\left( h\left( p_{k},\theta \right) \right) \right] _{L\times
1}\in \mathbb{R}^{L} \\
F_{q}\left( q \right) & =\left[ \frac{1}{N}\underset{k=1}{\overset{N}{%
\sum }}f_{i}\left( q_{k}\right) \right] _{L\times 1}\in \mathbb{R}^{L}.
\end{align}%

So far, the pose estimation problem has been transformed into an optimization
problem\ (\ref{optimization0})\ independent of correspondence, one key
characteristic of which is the separation of $p_{k}$ and $q_{k}$. 

\textbf{Remark 1}. The motivation of choosing $f_{i}$ is consistent with that in \cite{Hagege(2010), Domokos(2010), Domokos(2012)}. These independent
functions $f_{i}$ are called \emph{feature functions} here, which are expected to
describe the distinguishing features of the sets $\left \{ h\left(
p_{k},\theta \right) ,k=1,\cdots ,N\right \} $ and $\left \{
q_{k},k=1,\cdots ,N\right \} ,$ $i=1,\cdots ,L.$ This is consistent with the
requirement of feature extraction in pattern classification \cite{duda2001pattern}%
. Due to the space limit of the paper, we briefly discuss the mathematical principle of choosing feature functions that should follow the rule to let $\nabla f_{i}\ $\emph{orthogonal} on the set $\mathcal{Q},$ $i=1,\cdots,L.$ Besides, in order to avoid choosing feature functions case by case, the data and function from (\ref{mappingsim}) have to be normalized. We will focus on this in the future work.

\subsection{Practical Algorithm}

In this section, we will discuss some of the abnormal cases that we may encounter in practice.

\subsubsection{Unbalanced data sets}

An edge in 3D space is projected into two images which often consist of different numbers of
pixels, namely $M\neq N$, where $M=\left \vert \mathcal{P}\right \vert
,N=\left \vert \mathcal{Q}\right \vert .$ In this case, the method aforementioned cannot work. Before continuing, the following proposition is given without proof.

\textbf{Proposition 1}. The variables $q_{k}^{\prime }$ and $q_{k}^{\prime
\prime }$ are assumed to be independent and identically distributed random
variables with $\mathbb{E}\left( q_{k}^{\prime }\right) =\mathbb{E}\left(
q_{k}^{\prime \prime }\right) =\mu $ and cov$\left( q_{k}^{\prime }-\mu
\right) =$cov$\left( q_{k}^{\prime \prime }-\mu \right) =\Sigma _{0},$ $%
k=1,\cdots ,M$ $\left( \text{or }N\right) .$ Then%
\begin{align*}
\mathbb{E}\left( \frac{1}{M}\underset{k_{1}=1}{\overset{M}{\sum }}%
q_{k_{1}}^{\prime }-\frac{1}{N}\underset{k_{2}=1}{\overset{N}{\sum }}%
q_{k_{2}}^{\prime \prime }\right) & =0 \\
\text{cov}\left( \frac{1}{M}\underset{k_{1}=1}{\overset{M}{\sum }}%
q_{k_{1}}^{\prime }-\frac{1}{N}\underset{k_{2}=1}{\overset{N}{\sum }}%
q_{k_{2}}^{\prime \prime }\right) & =\left( \frac{1}{M}+\frac{1}{N}\right)
\Sigma _{0}.
\end{align*}

According to the \textbf{Proposition 1}, we have the following approximation
relation%
\begin{equation}
\frac{1}{M}\underset{k_{1}=1}{\overset{M}{\sum }}q_{k_{1}}^{\prime }\approx 
\frac{1}{N}\underset{k_{2}=1}{\overset{N}{\sum }}q_{k_{2}}^{\prime \prime }
\label{g}
\end{equation}%
This approach is particularly effective when both $M$ and $N$ are sufficiently large. Here, the concept of "large" is analogous to that in the "Law of Large Numbers" \cite{hsu1947complete}. A significant advantage of our correspondence-free method is that it enables the utilization of richer image features, such as edges, in addition to traditional corner points, without incurring substantial computational overhead that typically accompanies correspondence estimation. By (\ref{g}), we have $\frac{1}{M}\underset{k=1}{%
\overset{M}{\sum }}h\left( p_{k},\theta \right) $ $\approx \frac{1}{N}%
\underset{k=1}{\overset{N}{\sum }}q_{k}.$ Similarly, the parameter $\theta $
can  also be obtained by (\ref{optimization0}), where $F_{p}$ here is
slightly modified to be $F_{p}\left( p,\theta \right) =\left[ 
\begin{array}{c}
\frac{1}{M}\underset{k=1}{\overset{M}{\sum }}f_{i}\left( h\left(
p_{k},\theta \right) \right) 
\end{array}%
\right] _{L\times 1}\ $with $p=\left[ p_{k}\right] _{M\times 1}\in 
\mathbb{R}
^{n_{p}M}.$

When dealing with large values of $M$ and $N$, the average remains robust against a small number of \emph{outliers}. This requirement for large $M$ and $N$ is readily met, particularly in feature-rich environments. Additionally, we employ \textquotedblleft RANdom SAmple Consensus\textquotedblright \ (RANSAC), an iterative algorithm designed to estimate mathematical model parameters from observed data containing outliers \cite{1981Random}. By applying RANSAC for outlier detection, we further ensure that the point set adheres to our assumptions.

\subsubsection{Occlusions}

The occlusion in images is a problem that must be taken into account, especially in real scenes. 
Based on the principle of our algorithm, we only need to obtain a certain number of reference 
points with potential correct correspondence to optimize. Therefore, we introduce a search 
algorithm based on different grayscale images to deal with the presence of occlusion. Suppose the corresponding gray value of the $\mathcal{P}$ and $\mathcal{Q}$ are $G_P$ and $G_Q$ respectively. For these gray values, we apply K-means clustering to create two sets of clusters $S_P$ and $S_Q$ with size of $N$
\begin{align*}
    & S_P =\left\{ S^{P}_1,...,S^{P}_N \right \}, 
     S_Q =\left\{ S^{Q}_1,...,S^{Q}_N\right \}
\end{align*}

After obtaining these clusters, we calculate the distance between the mean gray values of every possible pair of clusters from $S_P$ and $S_Q$. Based on this distance metric, we select the top $N$ closest cluster pairs:

\begin{equation*}
    (S^{P}_{i_1},S^{Q}_{j_1}),...,(S^{P}_{i_N},S^{Q}_{j_N})
\end{equation*}

where $(S^{P}_{i_k},S^{Q}_{j_k})$ represents the cluster pair with the $k$-th smallest distance. The underlying mechanism is effective because occlusions in images typically appear in a scattered pattern and rarely dominate large continuous regions. These occlusions generally exhibit distinct gray values from the main features. Consequently, we can still obtain sufficient valid globally matched points. Furthermore, our correspondence-free approach offers an additional advantage: even in the presence of some mismatches, we can achieve accurate estimates by filtering out problematic subsets, thereby effectively mitigating the impact of occlusions.

\section{Simulation and Experiment}

\subsection{Settings and Problem Formulation}

At first, a pinhole camera with focal length $f_{c}$ is vertical to a plane with the latent depth $\lambda $ between the optical center. At the initial
position, we take an image of the reference points in the plane, where under
perspective projection these are denoted by $\left( P_{x,k},P_{y,k}\right) $%
, $k=1,\cdots ,N.$ After rotating by $R$ and translating by $T$, the camera
captures the second image. Their correspondences are denoted by $\left(
Q_{x,c\left( k\right) },Q_{y,c\left( k\right) }\right) $, where $c\left(
k\right) $ is an unknown correspondence function, $k=1,\cdots ,N.$ Alternatively, as the
depth $\lambda $ to be unknown, the objective is to determine $R$ and $\bar{T}%
=T\left/ \lambda \right. $. Let $P_{k}=[P_{x,k}\left/ f_{c}\right. $ $%
P_{y,k}\left/ f_{c}\right. $ $1]^{T}$ $\in $ $%
\mathbb{R}
^{3}$ and $Q_{c\left( k\right) }=[Q_{x,c\left( k\right) }\left/ f_{c}\right. 
$ $Q_{y,c\left( k\right) }\left/ f_{c}\right. $ $1]^{T}$ $\in $ $%
\mathbb{R}
^{3}.$ They have a relation that%
\begin{equation}
\frac{RP_{k}+\bar{T}}{\left \Vert RP_{k}+\bar{T}\right \Vert }=\frac{%
Q_{c\left( k\right) }}{\left \Vert Q_{c\left( k\right) }\right \Vert }.
\label{Relative}
\end{equation}

The rotation matrix $R=R\left( \alpha \right) $ has 3 Degrees of Freedom (3DoF)
with $\alpha \in \mathbb{R}^{3}$ being the Euler angles. The objective is to
estimate $\theta =\left[ 
\begin{array}{cc}
\alpha ^{T} & \bar{T}^{T}%
\end{array}%
\right] ^{T}.$ 

According to (\ref{Relative}) and to enhance robustness against noise, we normalize two sets in the form of (\ref{3dto2d}) as follows:
\begin{align}
\mathcal{P}& ={ p_{k}=\frac{\frac{R\left( \alpha \right) P_{k}+%
\bar{T}}{\left \Vert R\left( \alpha \right) P_{k}+\bar{T}\right \Vert }-\mu
}{2\sigma }\in
\mathbb{R}^{3},k=1,\cdots ,N } \label{Relative1}
\\
\mathcal{Q}& ={ q_{k}=\frac{\frac{Q_{k}}{\left \Vert Q_{k}\right \Vert }-\mu }{2\sigma }\in
\mathbb{R}^{3},k=1,\cdots ,N} .\label{Relative2}
\end{align}
where $\mu$ represents the mean value defined as $\mu =\frac{1}{N}\underset{k=1}{\overset{N}{\sum }}\frac{Q_{k}}{\left \Vert Q_{k}\right \Vert }$, and $\sigma$ denotes the standard deviation given by $\sigma =%
\sqrt{\frac{1}{N-1}\underset{k=1}{\overset{N}{\sum }}\left \Vert \frac{Q_{k}}{\left \Vert Q_{k}\right \Vert }-\mu
\right \Vert ^{2}}\in
\mathbb{R}.$

After normalization, each component of $q_{k}$ follows a normal distribution with probability density function $p\left(x\right)
=\frac{2}{\sqrt{2\pi}}e^{-2x^{2}}>0$. Notably, approximately 95

\subsection{Simulation}

\subsubsection{Data}

\begin{figure}[h]
\begin{center}
\includegraphics[
		scale=0.45]{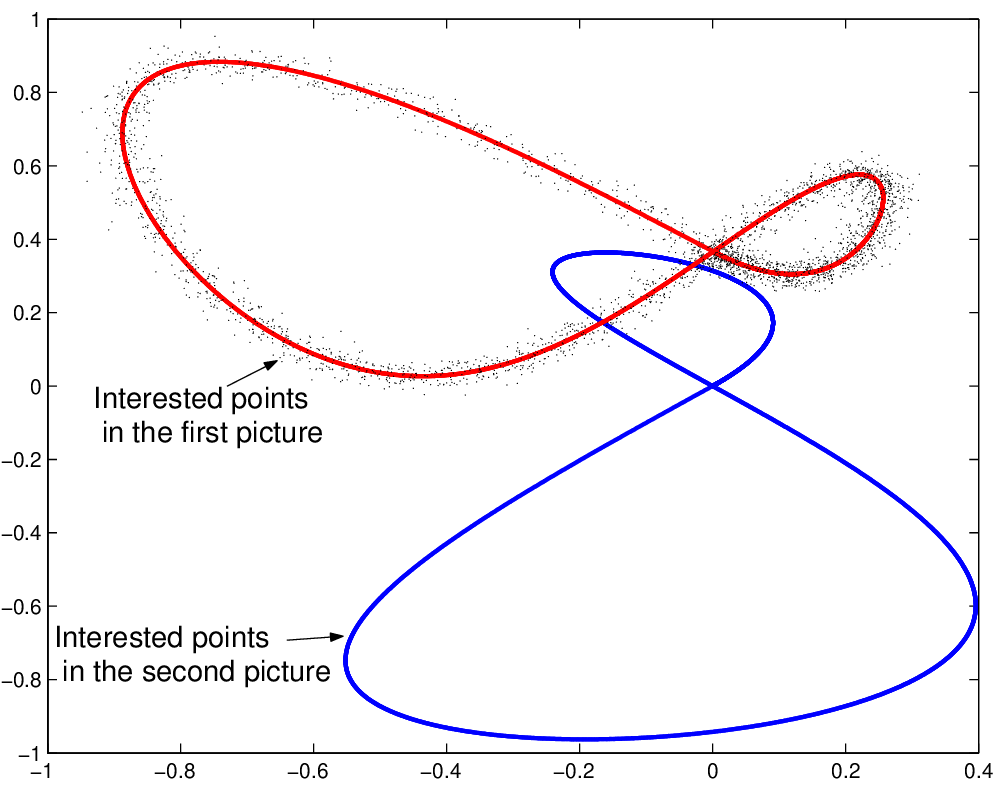}
\end{center}
\vspace*{-10pt}
\caption{Reference points (the blue and red curves) in two pictures, where
noisy reference points in the second picture with $b_{p}=0.02$ are 
represented in black points.}
\label{fig3}
\end{figure}

In the simulation, the data of the 3D-to-2D case is generated with a latent depth $\lambda$ of the first picture. The reference points in the two pictures are shown in Fig.~%
\ref{fig3} and they can be considered as the
component pixels of the smooth curve, for which it is not easy to extract
corners. In addition, we also introduce an adjustable term $b_p$ to add noise formulated as $b_{p}\text{randn}\left( 1,1\right)$ to the second picture.

\subsubsection{Feature Functions}

As explained in \textbf{Remark 1}, the intuition of choosing feature functions is to make them as different as
possible so that the optimal solution is insensitive to noise. In simulation, we chose $18$ feature functions
as follows%
\begin{align*}
f_{i}\left( x\right) & =-0.6578x_{i}-1.058\sin \left( \pi x_{i}\right)
+0.123x\cos \left( \pi x_{i}\right) \\
f_{3+i}\left( x\right) & =-0.2567x_{i}^{2}+0.3437\sin \left( \pi
x_{i}\right) ^{2}+1.286\cos \left( \pi x_{i}\right) \\
f_{6+i}\left( x\right) & =-0.7620x_{i}^{2}-1.288\sin \left( \pi x_{i}\right)
^{2}+0.1921\cos \left( \pi x_{i}\right) \\
f_{9+i}\left( x\right) & =1.245x_{i}-0.9539\sin \left( \pi x_{i}\right)
-1.540x\cos \left( \pi x_{i}\right) \\
f_{12+i}\left( x\right) & =2.998x_{i}^{2}-1.620\sin \left( \pi x_{i}\right)
^{2}+1.032\cos \left( \pi x_{i}\right) \\
f_{15+i}\left( x\right) & =-4.656x_{i}+2.290\sin \left( \pi x_{i}\right)
-5.183x\cos \left( \pi x_{i}\right)
\end{align*}%
where $i=1,2,3$ and the feature functions $\nabla f_{3(j-1)+i},j=1,...,6$ are orthogonal over the interval $(- \infty, +\infty)$ with respect to a weight function $p\left(  x\right)
=\frac{2}{\sqrt{2\pi}}e^{-2x^{2}}>0$. 
The feature functions are applied to each coordinate component of every point in sets $\mathcal{P}$ and $\mathcal{Q}$ independently. Specifically, each coordinate component for any point is processed through all feature functions. The selection of these feature functions adheres to the principles outlined in \textbf{Remark 1}, ensuring they effectively capture the underlying data characteristics of the normalized point sets defined in (\ref{Relative1}) and (\ref{Relative2}). This component-wise processing strategy enables comprehensive feature extraction while maintaining the geometric relationships inherent in the normalized data distribution. The initial condition is as follows:
\[
\theta_{0}=\theta^{\ast}+b_{i} randn\left(
6,1\right),
\]
where $b_{i}\in \mathbb{R}$.
\subsubsection{Results}

\begin{figure}[h]
\begin{center}
\includegraphics[scale=0.55]{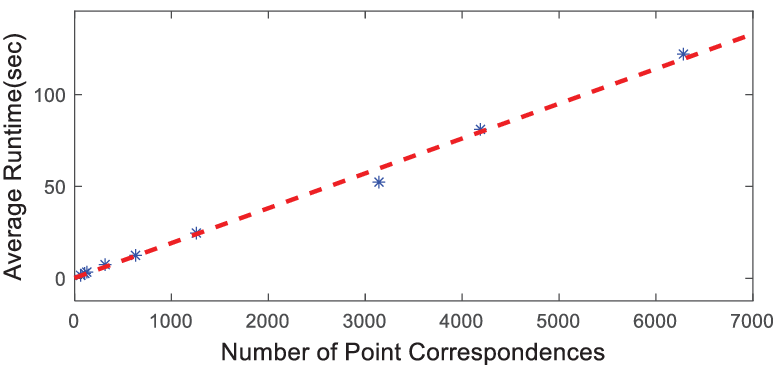}
\end{center}
\par
\caption{The relation between the average runtime and the number of points
correspondence.}
\label{runtim}
\end{figure}

To show the efficiency of the proposed algorithm, we implement the
optimization problem (\ref{optimization0}) with different numbers of
correspondence points \footnote{%
The computation is performed by MATLAB R2017b on a desktop computer (DELL
v3881) with Intel Core i7-10700 CPU at 2.90 GHz.} when $b_{i}=0.2$ and $%
b_{p}=0$. As shown in Fig.~\ref{runtim}, the average runtime increases
approximately linearly as the number of points correspondence increases. 

Next, the performance of the proposed algorithm in the presence
of noise, mismatches, and outliers is shown in the pictures. i) \textbf{Noise without outliers}. Perform $100$
trials with different $\left( b_{p},b_{i}\right) $ randomly and let $\hat{\theta%
}$ be the estimate. If $|| \hat{\theta}-\theta^{\ast } || \leq0.1,$ then the
trial is considered to be successful. Otherwise, the trial fails. Finally,
we have the results shown in Table~\ref{table:headings1}. (a). In the presence of
appropriate noise, we can still find the approximated
parameter with a high possibility. ii) \textbf{Mismatches without outliers}. Suppose that the pairs $\left(
Q_{x,k},Q_{y,k}\right) $ are selected for the set $\mathcal{Q}$ if $%
\left
\vert \text{randn}\left( 1,1\right) \right \vert \leq
b_{m},k=1,\cdots,3142.$ As a result, the numbers of elements of $\mathcal{P}$
and $\mathcal{Q}$ are different with $M=3142$ in the first picture and only $%
N=1239,2149,2734$ in the second picture if $b_{m}=0.5,1,1.5,$ respectively.
Perform $100$ trials for different $\left( b_{i},b_{m}\right) $ randomly.
As shown in Table~\ref{table:headings1}. (b), we still have a certain possibility
of finding the approximated parameter in the presence of noise and
mismatches. iii) \textbf{Outliers}. Set $M=N=3124,b_{p}=0.02$ at first. Then, add 150
outliers in the second picture, which are at $\left( -0.6+0.05\text{rand}%
\left( 1,1\right) ,-0.4+0.05\text{rand}\left( 1,1\right) \right) $ randomly.
Initial Condition:\textbf{\ }$\theta
_{0}=\theta ^{\ast }+0.2\ast $rand$\left( 6,1\right) .$ At first, parameter $\hat{%
	\theta}_{1}$ with $\left\Vert \hat{\theta}_{1}-\theta ^{\ast }\right\Vert
=0.4129$ is obtained. After using RANSAC, we remove the outliers and further obtain $%
\hat{\theta}_{2}$ with $\left\Vert \hat{\theta}_{2}-\theta ^{\ast
}\right\Vert =0.0577.$ It is easy to see that RANSAC helps to get a better result.%
\setlength{\tabcolsep}{4pt} 
\vspace{-0.2cm}
\begin{table}[h]
\caption{Success number in 100 trials}
\label{table:headings1}
\begin{center}
\begin{tabular}{|c|c|ccc|}
\hline
\multirow{3}*{(a)}
 & $b_{i}\left \backslash b_{p}\right. $ & 0.01 & 0.02 & 0.03 \\ \cline{2-5}
 & 0.1 & 100 & 100 & 69 \\ 
 & 0.2 & 99 & 98 & 52 \\ \hline 
\multirowcell{3}{(b)}
 & $b_{i}\left \backslash b_{m}\right. $ & 0.5 & 1 & 1.5 \\ \cline{2-5}
 & 0.1 & 38 & 85 & 98 \\ 
 & 0.2 & 36 & 86 & 95 \\ \hline
\end{tabular}%
\end{center}
\end{table}
\vspace{-0.4cm}
\subsubsection{Comparison}
\begin{figure}[h]
\begin{center}
\includegraphics[
		scale=0.8]{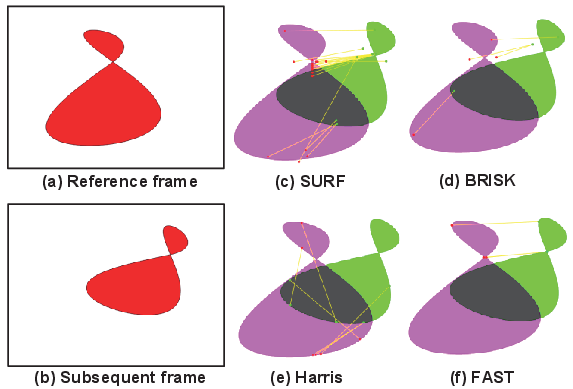}
\end{center}
\par
\vspace*{-5pt}
\caption{The match results through different feature extraction methods. (a)
and (b) are the reference and subsequent frames which represent the images
taken by a camera at different positions. The purple and green points in
(c)-(f) are the extracted feature points based on different common feature
extraction methods. The corresponding relationship between them is indicated
by yellow lines.}
\label{figcom}
\end{figure}

\begin{figure}[h]
\vspace{-0.4cm}
\begin{center}
\includegraphics[
		scale=0.85]{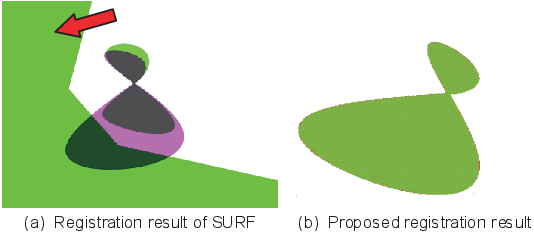}
\end{center}
\par
\vspace*{-10pt}
\caption{The registration results. Here (a) and Fig.~\ref{figcom} (c)-(f) are generated by the MATLAB R2017b toolbox, Registration Estimator. There is a wrong registration in (a) and the green area indicated by the red arrow is the content beyond the field of the subsequent frame, just for the purpose of distinguishing. The green color shape has overlapped the purple one completely in (b).}
\label{register}
\vspace{-0.5cm}
\end{figure}

We plot the reference points and
fill the interior with red color for visualization as shown in Fig.~\ref%
{figcom}(a). A subsequent frame is shown in Fig.~\ref{figcom}(b) after
rotating and translating the camera (purple and green are used to distinguish
the two images in Fig.~\ref{figcom}(c)-(f)). Fig.~\ref{figcom} (c)-(f) are
the results of feature extraction through different feature extraction
methods such as SURF\cite{surf}, BRISK\cite{BRISK(2011)}, Harris\cite%
{Harris(1988)} and FAST\cite{FAST(2006)} . From Fig.~\ref{figcom} (c)-(f),
obviously, these methods mentioned above are inapplicable because there are
few correspondent feature point pairs or existing many mismatches which may
lead to a bad pose estimation result, as shown in Fig.~\ref{register}(a) by
taking SURF for example. On the other hand, the proposed correspondence-free
pose estimation method gets a good matching result shown in Fig.~\ref%
{register}(b). This implies that a correct pose is obtained. 

\subsection{Experiment}
In this subsection, we verify the proposed pose estimation
algorithm in two cases, i.e. 3D-to-2D and 2D-to-2D.

\subsubsection{3D-to-2D scenario}
In this case, we performed three different experiments based on the relation \ref{3dto2d} for different scenes and different numbers of unknowns. Firstly, we prepare a mobile robot with a camera and print a pattern on white paper attached to a planar wall. At the initial position, the optical axis of the camera is vertical to the wall. Then, a picture with the given pattern is taken as a reference. After that, move the robot to take pictures in different poses. The patterns are within the camera's field of view all the time. Next, the red patterns in all pictures (a series of video frames) are segmented simply by a threshold in HSV space. Since the depths are different, the numbers of pattern pixels in different pictures are different. Finally, the estimate is obtained by solving the optimization problem (\ref{optimization0}) directly. By using the obtained pose, we align the interested points in different pictures with those in the reference picture,  which is called registration. The results are shown in Fig.~\ref{exper} (for more, please refer
to the video in supplemental materials) which implies that the estimated pose is close to the ground truth.

\begin{figure}[h]
\vspace{-0.3cm}
\begin{center}
\includegraphics[
		scale=0.75]{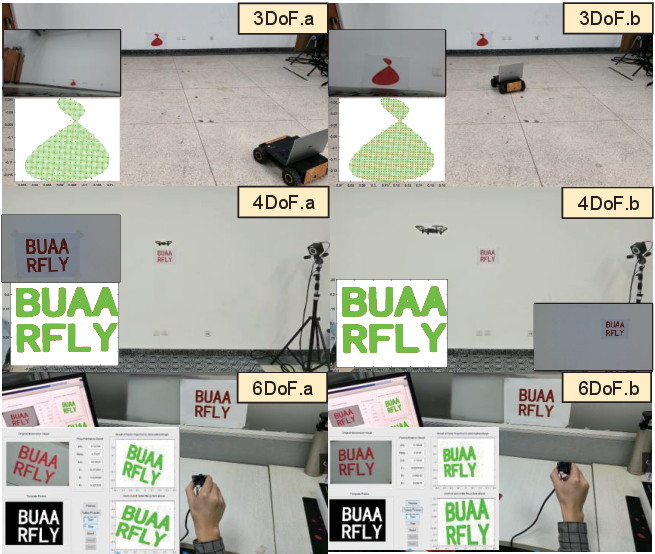}
\end{center}
\vspace*{-10pt}
\caption{The results of three indoor experiments. Different experiments solve different numbers of unknowns. There is a mobile robot, a flying aerial robot, and a camera that can move freely which represents 3DoF (yaw and 2D position x, y), 4DoF (yaw and 3D position x, y, z), and 6DoF (pitch, roll, yaw, and 3D position x, y, z) experiments,  respectively. The camera's view, patterns, and registration results are all shown in each figure.}
\label{exper}
\vspace{-0.25cm}
\end{figure}

\begin{figure}[H]
\vspace{-0.3cm}
\hspace{0.2cm}
\includegraphics[width=8.1cm]{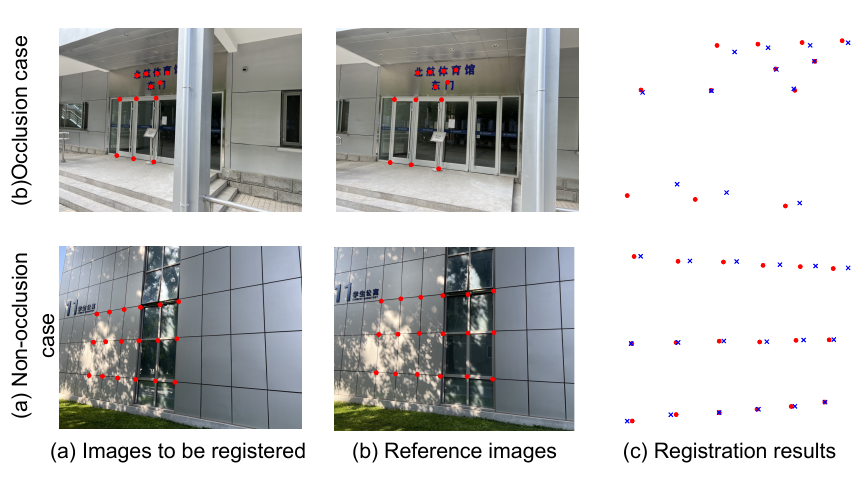}
\caption{The results of outdoor experiments. 6DoF, 2D-to-2D registration under general lighting conditions. The left side is the images to be registered, and the middle is the reference images. The right side shows the registration results.}
\vspace{-0.3cm}
\label{2d-2d rlt}
\end{figure}
\subsubsection{2D-to-2D scenario}
In this case, based on relation \ref{2dto2d}, we conduct outdoor 6DoF experiments with flexible configurations to simulate real-world scenarios that a robot might encounter, as illustrated in Fig.\ref{2d-2d rlt}. To evaluate the accuracy of our method under both occlusion and non-occlusion conditions, we capture two sets of outdoor images using an iPhone 13 Pro camera and manually annotate the corresponding key points in these image pairs.

For evaluation, we draw corresponding red dots artificially as shown in these images to check the accuracy of the method. We will reproject the points in the left photos based on the obtained $R$ and $t$ and compare them with the points in the right photos. The points are expected to be as close to the true points as possible. 

In Fig.~\ref{2d-2d rlt}, the red dots indicate the ground truth locations of the reference points, while the blue markers denote the estimated positions derived from our correspondence-free method applied to the left images in 2D-2D cases. For images containing occlusions, we implemented the previously described approach to minimize the impact of occlusions on feature point extraction. Although the accuracy slightly decreases compared to non-occluded scenarios, the results remain within acceptable bounds.

\section{Conclusion}

In this work, we have introduced a novel correspondence-free approach for pose estimation. The primary advantage of our method over existing correspondence-free techniques lies in its broader applicability, particularly in handling nonlinear perspective projection cases. This capability enables our approach to address virtually all pose estimation scenarios encountered in computer vision applications. Nevertheless, our research has certain limitations. A notable challenge remains in the optimal selection of feature functions - specifically, identifying functions that either enhance the convexity of the optimization problem to facilitate global minimum convergence or improve robustness against noise and mismatches. In future work, we plan to investigate adaptive feature functions integrated with modern embedding techniques in computer vision. The enhanced method will be rigorously evaluated against state-of-the-art correspondence-based approaches to demonstrate its effectiveness.

\bibliographystyle{IEEEtran}
\bibliography{root}

\end{document}